%% file: main.tex
\documentclass[10pt,twocolumn,letterpaper]{article}

\usepackage[pagenumbers]{cvpr} 

\usepackage{tabularx}
\usepackage{subcaption}
\usepackage{booktabs}
\usepackage{float}
\usepackage{url}
\usepackage{placeins}
\usepackage{mwe}
\usepackage{graphicx}
\usepackage{amsmath, amssymb, mathtools}
\usepackage{adjustbox}
\usepackage{hhline}

\newcommand{\cN}{\mathcal{N}}
\newcommand{\cP}{\mathcal{P}}
\newcommand{\bbE}{\mathbb{E}}

\input{preamble}

\definecolor{cvprblue}{rgb}{0.21,0.49,0.74}
\usepackage[pagebackref,breaklinks,colorlinks,citecolor=cvprblue]{hyperref}

\title{
Improving In-Context Learning in Diffusion Models\\ with Visual Context-Modulated Prompts
}

\author{Tianqi Chen$^1$\quad Yongfei Liu$^2$\quad Zhendong Wang$^1$\quad Jianbo Yuan$^2$\quad Quanzeng You$^2$\\
Hongxia Yang$^2$\quad Mingyuan Zhou$^1$\\
\textsuperscript{1}The University of Texas at Austin\quad\textsuperscript{2} ByteDance Inc.\\
{\tt\small \{tqch,zhendong.wang\}@utexas.edu}\\
{\tt\small \{liuyongfei.0314,jianbo.yuan,quanzeng.you,hx.yang\}bytedance.com}\\
{\tt\small mingyuan.zhou@mccombs.utexas.edu}
}

\begin{document}

\makeatletter
\let\@oldmaketitle\@maketitle%
\renewcommand{\@maketitle}{\@oldmaketitle%
    \centering
    \includegraphics[width=0.9\linewidth]{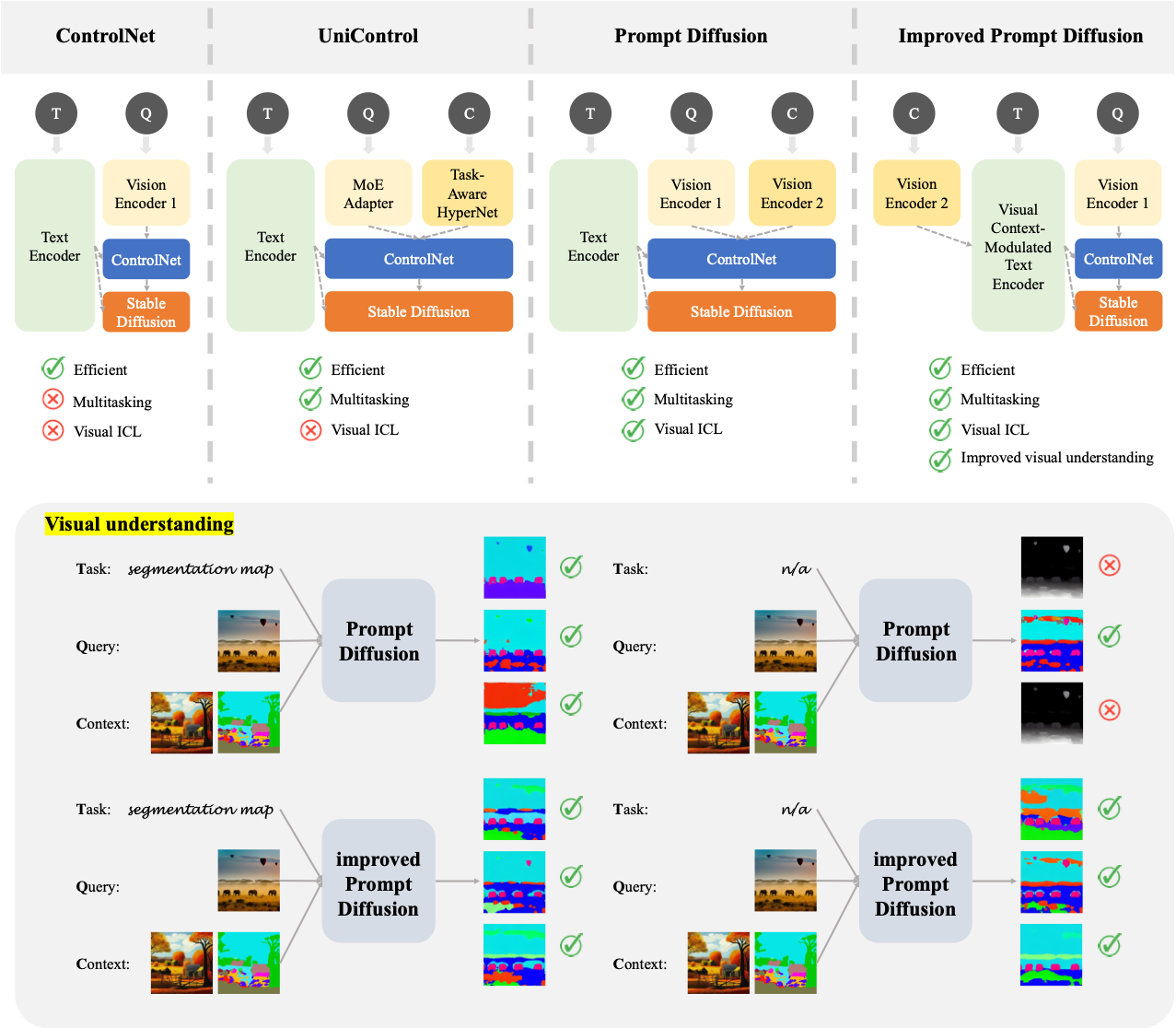}
    \captionof{figure}{\textbf{Comparative Analysis of Diffusion Models in Visual In-Context Learning.} This figure showcases a comparative review of four diffusion-based methods, emphasizing their unique architectural designs, alongside their respective strengths and limitations in executing vision tasks. Our model, iPromptDiff, which builds upon and enhances the original Prompt Diffusion framework, is specifically engineered to improve visual understanding. The figure demonstrates the advancements of our model through example comparisons in the image-to-segmentation task, showcasing its performance in scenarios both with and without an explicit textual task description.}
    \label{fig:teaser}
   \bigskip}         
\makeatother

\maketitle

\input{sec/0_abstract}    
\input{sec/1_intro}
\input{sec/2_related_work}
\input{sec/3_method}

\input{sec/4_exp}
\input{sec/5_discussion}
\input{sec/6_conclusion}


{
    \small
    \bibliographystyle{ieeenat_fullname}
    \bibliography{main}
}

\appendix

\input{sec/X_suppl}

\end{document}

%% file: preamble.tex
\usepackage[dvipsnames]{xcolor}


%% file: sec/0_abstract.tex
\begin{abstract}

In light of the remarkable success of in-context learning in large language models, its potential extension to the vision domain, particularly with visual foundation models like Stable Diffusion, has sparked considerable interest. Existing approaches in visual in-context learning frequently face hurdles such as expensive pretraining, limiting frameworks, inadequate visual comprehension, and limited adaptability to new tasks. In response to these challenges, we introduce improved Prompt Diffusion (iPromptDiff) in this study. iPromptDiff integrates an end-to-end trained vision encoder that converts visual context into an embedding vector. This vector is subsequently used to modulate the token embeddings of text prompts. We show that a diffusion-based vision foundation model, when equipped with this visual context-modulated text guidance and a standard ControlNet structure, exhibits versatility and robustness across a variety of training tasks and excels in in-context learning for novel vision tasks, such as normal-to-image or image-to-line transformations. The effectiveness of these capabilities relies heavily on a deep visual understanding, which is achieved through relevant visual demonstrations processed by our proposed in-context learning architecture.

\end{abstract}

%% file: sec/1_intro.tex
\section{Introduction}
\label{sec:intro}

In-context learning has made significant strides in the language domain, largely thanks to advancements in large language models \cite{brown2020language, ouyang2022training, chowdhery2022palm}. While this progress is evident in natural language processing (NLP), the exploration of in-context learning capabilities within large vision foundation models is still in its early stages. We can divide the existing approaches into two categories: inpainting-based methods~\cite{bar2022visual, Wang_2023_CVPR, Wang_2023_ICCV} and diffusion-based methods~\cite{wang2023incontext}. These methods, when capable of in-context learning, often encounter issues such as costly pretraining, restrictive problem formulation, limited visual understanding, and reduced effectiveness in out-of-domain (OOD) scenarios. For example, masked-image-modeling (MIM) techniques like Painter \cite{Wang_2023_CVPR} and SegGPT \cite{Wang_2023_ICCV} necessitate training from scratch, and they also demand specific input structure ($e.g.$, a two-by-two image grid) of in-context examples and queries due to their in-painting problem formulations.

Conversely, Prompt Diffusion \cite{wang2023incontext} presents an efficient, finetuning-based alternative. This method effectively utilizes the ControlNet \cite{Zhang_2023_ICCV} within its controllable generation process to accommodate various visual in-context examples and to facilitate diverse visual tasks. However, it's important to acknowledge that Prompt Diffusion can be overly sensitive to text prompts, as our research indicates. Also, Prompt Diffusion has an intertwined visual input setting, $i.e.$, both the image query (low-level image semantics) and the in-context examples (high-level task semantics) are encoded through vanilla-CNN-based encoders in the same way and the representations are directly summed as one single hint vector for the ControlNet. This tendency towards overfitting could negatively impact its efficacy in new vision tasks that demand sophisticated visual understanding.

Considering these challenges, we introduce an enhanced version of Prompt Diffusion, termed Improved Prompt Diffusion (iPromptDiff), designed to address the four key issues: expensive pretraining, restrictive problem formulation, limited visual understanding, and insufficient OOD generalizability. At the heart of our method is a vision encoder-modulated text encoder. This system skillfully extracts high-level visual context from a pair of example images and transforms it into a visual embedding vector. This vector is then used to modulate the token embeddings of the text prompts, thereby enhancing their contextual relevance. These enhanced embeddings are subsequently processed by the ControlNet and Stable Diffusion components, enabling them to perform image generations that are richly contextualized and aligned with the provided visual cues.

Our main contributions are as follows:
\begin{enumerate}
\item We present a unique strategy to enhance visual comprehension in visual in-context learning. This is achieved by uncoupling the processing of visual context from that of image queries, and then effectively integrating this context to modulate the textual input.

\item We develop a vision encoder that can be seamlessly plugged into existing text-to-image diffusion models. This encoder acts as a conduit for an additional visual signal, providing crucial visual in-context information.

\item We carry out comprehensive qualitative and quantitative assessments to evaluate the efficacy and generalizability of our proposed method. Our findings demonstrate its performance to be 
superior to existing state-of-the-art models. Notably, it exhibits a reduced tendency for text overfitting.
\end{enumerate}

%% file: sec/2_related_work.tex
\section{Related Work}

\subsection{Diffusion Models}

Gaussian diffusion models, initially introduced by \citet{sohl2015deep} and further popularized by  \citet{song2019generative}, \citet{ho2020denoising}, and \citet{song2020score}, have become increasingly prominent in deep generative modeling \cite{dhariwal2021diffusion, karras2022elucidating, ramesh2022hierarchical}. Spurred by their success, researchers have developed other diffusion model variants for a wider array of data types, including categorical data \cite{hoogeboom2021argmax, austin2021structured}, count data \cite{chen2023learning}, and range-bounded continuous data \cite{zhou2023beta}. Despite these advancements, Gaussian diffusion models continue to be the most influential in image generation. A notable example is Latent Diffusion \cite{rombach2022high}, which has been instrumental in advancing high-quality image generation and stimulating the growth of diffusion-based text-to-image vision foundation models \cite{saharia2022photorealistic, balaji2022ediffi, peebles2023scalable}.

\subsection{Visual In-Context Learning}

In-context learning is a paradigm extensively used in NLP to elicit models' adaptability to unfamiliar tasks like translation, question-answering, and complex reasoning \cite{brown2020language, min-etal-2022-metaicl, wei2022finetuned, wei2022chain}. In NLP, models can tackle unseen tasks using in-context examples, such as pairs of text and labels. However, visual in-context learning in the vision domain presents more significant challenges and remains under-explored. 

One important challenge is that, unlike language models which process variable-length text inputs, vision models typically cannot easily handle inputs of arbitrary lengths. It is often not feasible to simply concatenate numerous images into a single input for standard vision foundation models. Another critical issue is the nuanced and intricate visual understanding required. When a task's specifics are not directly stated and must be inferred from a few image examples, it becomes particularly challenging for vision models to discern and comprehend the high-level visual relationships present in these examples.

Recently, the application of masked image modeling has been pivotal in enhancing the in-context learning capabilities of vision models \cite{bar2022visual, Wang_2023_CVPR, Wang_2023_ICCV}. The approach by \citet{bar2022visual} utilizes a masked-autoencoder-based (MAE) method. This method involves predicting a missing image in a two-by-two image grid, where two images serve as in-context examples and another image functions as the query. Subsequently, \citet{Wang_2023_CVPR,Wang_2023_ICCV} have expanded this concept to a multitask framework. Despite their advancements, such inpainting methods inherently require a fixed number of in-context examples (in this case, only one example in the two-by-two grid) and result in increased memory usage. Painter, as described in \citet{Wang_2023_CVPR}, stands out as a notable example of an inpainting-based method. It is crafted to function as a versatile tool applicable across a range of vision tasks.

Contrasting with the inpainting-based approach for visual in-context learning, \citet{wang2023incontext} modified the ControlNet framework, as initially presented in \citet{Zhang_2023_ICCV}. Their adaptation involved integrating an extra pair of example images and training the model through a multitask supervised finetuning (SFT) technique. This modification gave rise to Prompt Diffusion, a model showing notable proficiency in visual in-context learning. Nonetheless, their findings also reveal scenarios where the model's performance might not meet expectations, underscoring some constraints in its practical use.

\subsection{Image-to-Image Translation}
Image-to-image (I2I) translation is a task that involves maintaining the intrinsic content of a source image while adopting the extrinsic style of a target image~\cite{pang2021image}. This concept has been explored in various generative modeling frameworks, such as GANs~\cite{isola2017image, yi2017dualgan, liu2017unsupervised, lin2018conditional, choi2018stargan} and diffusion models~\cite{kim2022diffusionclip, peng2023diffusion, tumanyan2023plug}. In the realm of visual in-context learning, paired images that showcase specific image tasks, like visual in-context examples, are also considered forms of image-to-image translation.

%% file: sec/3_method.tex
\section{Improved Prompt Diffusion}
\label{sec:ipromptdiff}

\begin{figure}[!t]
    \centering
    \includegraphics[width=0.45\textwidth]{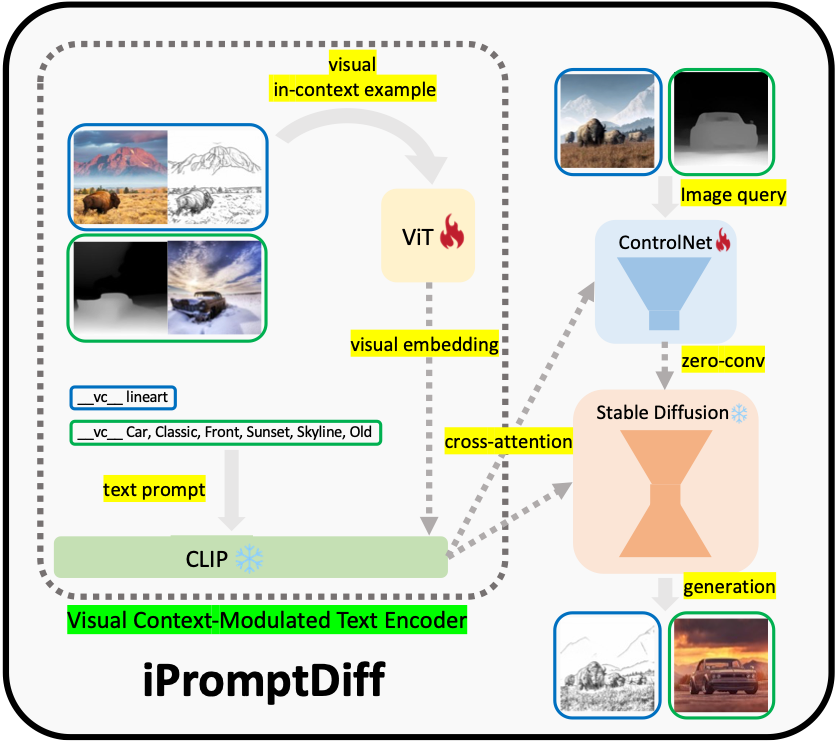}
\caption{\textbf{Configuration of Improved Prompt Diffusion (iPromptDiff).} This diagram illustrates the operational framework of the iPromptDiff model. It shows the model's processing flow, where it takes a pair of images as the visual context and an image query as inputs, along with an optional text prompt, to produce an output image. To aid understanding, two example sets of input-output pairs are highlighted in distinct color-coded boxes (\textcolor{blue}{blue} and \textcolor{green}{green}). The left section of the diagram represents a visual context-modulated text encoder. The output from this encoder is then cross-attended by the components on the right, facilitating contextualized generation. External links between the inputs/outputs and the iPromptDiff modules are indicated by solid arrows, while the internal connections within the iPromptDiff system are shown with dotted arrows. 
    \label{fig:model_arch}}
\end{figure}

In this study, we introduce the ``improved prompt diffusion'' (iPromptDiff) method, which integrates four key components: a CLIP \cite{radford2021learning} text encoder $E_\psi^{(T)}$, a vision-transformer-based \cite{dosovitskiy2021an} vision encoder $E_\phi^{(I)}$, Stable Diffusion \citep{rombach2022high} $S_\theta$, and ControlNet \citep{Zhang_2023_ICCV} $C_{\theta'}$.
 The architecture of the overall model is illustrated in Figure~\ref{fig:model_arch}. 
 Both the CLIP and Stable Diffusion components are frozen to keep the ability of the pretrained vision and language foundation models intact.
 The model operates with $x$ representing the data, $z_t$ the latent state at timestep $t$, $\epsilon_t$ the noise at timestep $t$, $p$ the text prompt, and $v$ the visual in-context example. Assuming the underlying task, denoted by $c$, follows a task distribution $\cP_C$,
 the training objective of iPromptDiff can be viewed as a generalization of the simple denoising loss~\cite{ho2020denoising} of diffusion models:

\begin{align}
   \min_{\theta',\phi} \bbE_{c\sim\cP_C}\bbE_{z_t,t,p,v,\epsilon_t\sim\cN(0,I)}\|\epsilon_t-\epsilon_{\theta'}(z_t,t, e_\phi(p, v))\|_2^2 \notag
\end{align}

\subsection{Vision Encoder}\label{subsec:vision_enc}
A vision encoder is one of the vital parts in the visual understanding and in-context learning process. It plays an important role in perceiving the relational similarity between the example images and informing the downstream generative modules of the high-level context (task) information. Looking into the design of Prompt Diffusion, we identify an entwined interface for both image query and visual in-context examples, $i.e.$, both of them are encoded through CNN-based VAE encoders and sent to the ControlNet module after elementwise arithmetic addition of their hidden representations. This may hamper the visual understanding of the input conditions, as the two reflect distinct levels of visual content (low-level image semantics versus high-level task semantics). Therefore, we choose a more capable vision model than the CNN-based VAE encoder used by Prompt Diffusion. 

Specifically, to achieve better visual understanding, we employ a vision transformer (ViT) with a modified number of input channels as our vision encoder to extract context embeddings from visual example pairs and account for the intrinsic relationship between the example images, which is a pair of images concatenated along the channel dimension in this case. 
To maintain simplicity in our approach, we chose not to engage in any architecture search. Instead, we consistently used the ViT-L/14 model configuration (with the sole exception of the input convolution layer) in all our experiments. Throughout the training process, the vision encoder was trained end-to-end in conjunction with the ControlNet.

\subsection{Visual Context Token} \label{subsec:vc_token}
A naive way to leverage the extracted visual embeddings would be to concatenate the visual embeddings directly to the CLIP text embeddings. However, we argue that this approach may be suboptimal and can undermine the model performance in certain cases because of missing interaction between vision and language context and potential information conflicts. 

Instead of simple concatenation, we propose to prepend a special visual context placeholder, \text{``\_\_vc\_\_''} in our case, to the text prompt. The special visual context placeholder is constructed and reserved such that it can be uniquely mapped to a new token of the vocabulary without interfering with the original text tokens, whose embedding will later be replaced with the vision context extracted by the vision encoder. Intuitively, learning visual context embeddings and fusing vision context with text guidance in this way can encourage a harmonic synergy between the two modalities and alleviate information conflicts. This approach also shares a similar spirit with textural inversion~\cite{gal2022image}, which has been an effective way of conceptualizing any vision task and implanting it to the language model. The key difference, however, lies in the flexibility of our method. Since the embedding of our special token is a function in terms of visual in-context examples and is always inferred dynamically depending on the context input, it is much more flexible than a single learnable token embedding by textual inversion.

\begin{figure}[t]
    \centering
    \includegraphics[width=0.45\textwidth]{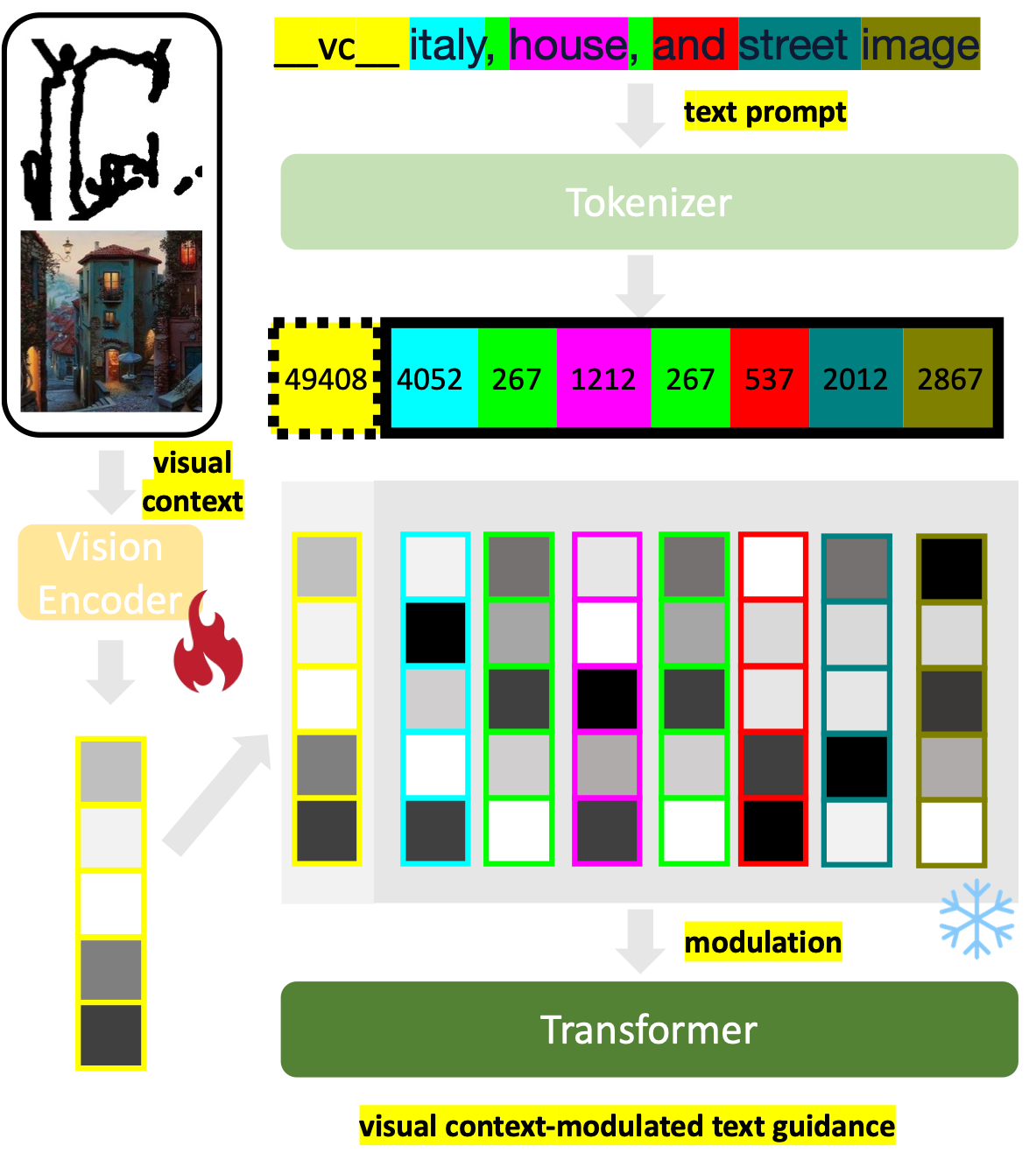}
    \caption{\textbf{Methodology for Creating Visual Context-Modulated Text Prompts.} The diagram illustrates the process of generating text prompts that are modulated by visual context. The vision encoder produces a visual embedding vector that holds high-level visual context information. This vector is tailored to substitute the embedding of a designated special token ($i.e.$, \text{``\_\_vc\_\_''}) in the text prompt.
    Following this substitution, the CLIP transformer employs the visual embedding vector to expertly modulate the text embeddings with the visual context.
    }
\end{figure}

\subsection{Multitask Training}
Echoing the approach of earlier vision generalist models~\cite{Wang_2023_CVPR, wang2023incontext}, we train iPromptDiff end-to-end with a multitask objective. We anticipate that by exposing the model to a diverse range of vision tasks, both the vision encoder and the extracted visual context embedding will develop enhanced robustness against variability in visual in-context examples and text prompts. This robustness, we believe, will enhance the model's ability to effectively generalize to unseen tasks it has not previously encountered.

%% file: sec/4_exp.tex
\section{Experiments}
\label{sec:exp}

In this section, we provide a comprehensive empirical evaluation of our proposed iPromptDiff model, focusing on both its in-domain and out-of-domain performance. For map-to-image tasks, which entail controllable generation to natrual images, iPromptDiff achieves results comparable to existing baseline models in in-domain settings. However, it exhibits distinct advantages in out-of-domain scenarios. This enhanced adaptability to previously unseen, out-of-domain tasks is similarly evident in the image-to-map tasks, further showcasing the model's robust generalizability.

\textbf{Implementation. } Our codebase is developed based on the official Prompt Diffusion implementation\footnote{\url{https://github.com/Zhendong-Wang/Prompt-Diffusion}}, which itself is an extension of the ControlNet codebase\footnote{\url{https://github.com/lllyasviel/ControlNet}} and the Instruct-Pix2Pix codebase\footnote{\url{https://github.com/timothybrooks/instruct-pix2pix}}. The vision encoder within our framework is adapted from the ViT code available in PyTorch\footnote{\url{https://github.com/pytorch/vision/blob/main/torchvision/models/vision_transformer.py}}. For additional information and specifics about our implementation, please refer to the Appendix.

\textbf{Datasets. }
In our experiments, we utilized two datasets. Firstly, for a direct comparison, we trained our model using the same dataset employed by Prompt Diffusion, namely, the CLIP-filtered Instruct Pixel-to-Pixel dataset. This dataset is a Stable Diffusion synthetized image-edit-prompt dataset comprising of 313,010 image editing pairs. All the condition maps were generated with the annotators implemented in ControlNet.

Additionally, we trained our model on a more extensive dataset, MultiGen-20M, which was curated by \citet{qin2023unicontrol}. The MultiGen-20M dataset encompasses around 20 million high-quality real image-map pairs, covering a spectrum of 12 different tasks. For both datasets, we allocated five percent of the total data as a test set, which was used for evaluation purposes.

\textbf{Evaluation. }
We provide quantitative evaluation results of our model in two different types of tasks including map-to-image and image-to-map. The evaluation protocol follows the one used by Prompt Diffusion, where reconstruction FIDs are reported for map-to-image tasks and RMSEs are reported for image-to-map tasks. Specifically, we approximate the ground truth condition maps with pretrained annotators and generate 10,000 images in total to compute these evaluation metrics against the test set.

\begin{table}[t!]
\centering
    \begin{adjustbox}{width=\linewidth}
    \begin{tabular}{@{} lccc @{}}
        \toprule
        Model & Depth-to-Img & HED-to-Img & Seg-to-Img \\
        \midrule
        ControlNet & 19.81 & 13.07 & 20.71 \\
        PromptDiff & 18.60 & 13.35 & 19.46 \\
        iPromptDiff (ours) & 16.47 & 11.24 & \textbf{17.32} \\
        iPromptDiff-MG20M (ours) & \textbf{11.21} & \textbf{9.02} & 24.03 \\
        \bottomrule
    \end{tabular}
    \end{adjustbox}
    \caption{\textbf{FID Comparison:} This table illustrates the FID scores calculated between test set samples and their corresponding reconstructions by different models on the Instruct-Pix2Pix dataset. }
    \label{tab:fid}
\end{table}

\begin{figure*}[h!]
    \centering
    \includegraphics[width=0.9\textwidth]{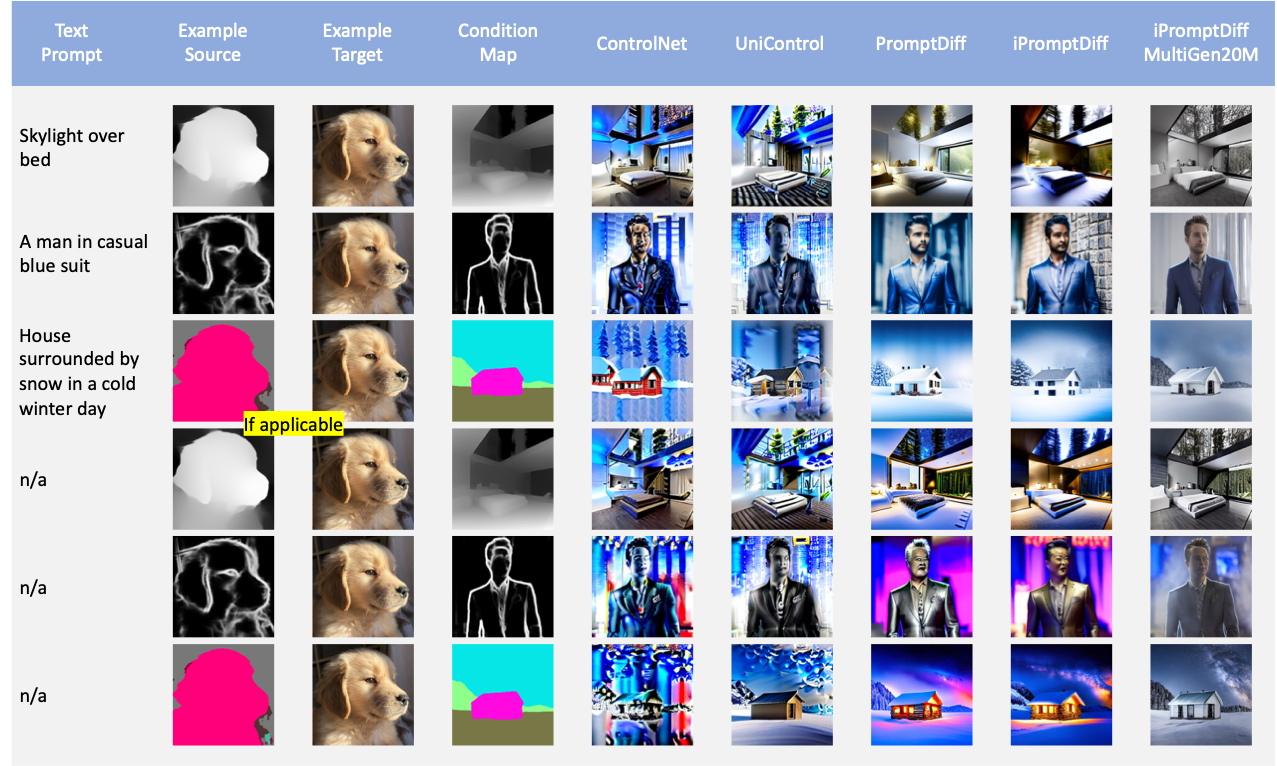}
    \caption{\textbf{Comparison of Controllable Generation Results for Previously Seen, In-Domain Tasks:} This figure displays the outcomes from ControlNet, UniControl, Prompt Diffusion, our improved Prompt Diffusion, and our improved Prompt Diffusion trained on MultiGen20M. The first column shows the text prompt, with `n/a' indicating the absence of a text prompt. The second and third columns showcase the source and target images from the provided visual context, which are not utilized by ControlNet and UniControl. The fourth column contains the query image, while the subsequent columns present the generated results using the various methods mentioned.}
    \label{fig:map2img}
\end{figure*}

\begin{figure}[h!]
    \centering
    \includegraphics[width=0.45\textwidth]{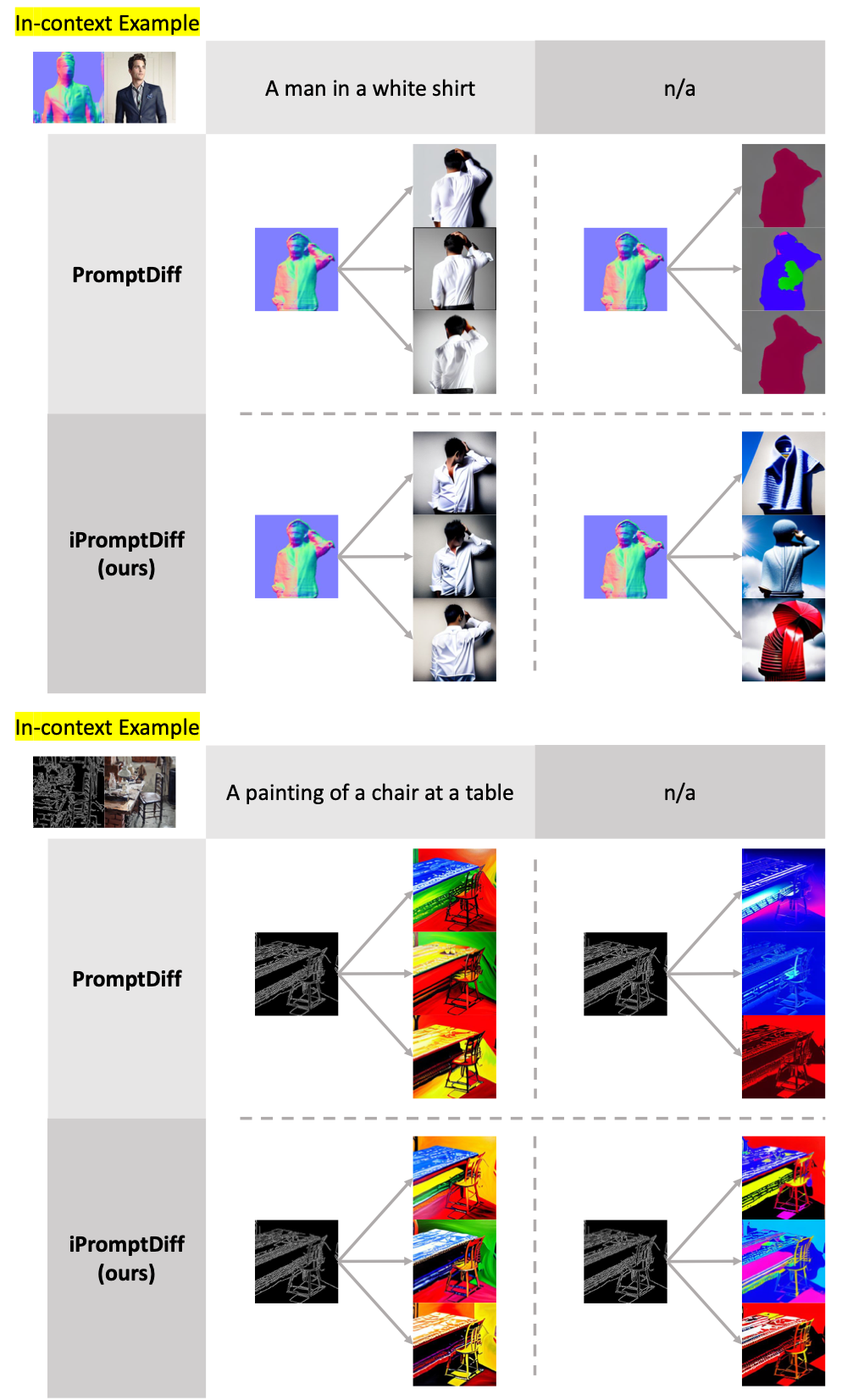}
    \caption{\textbf{Comparison of Unseen Task Performance:} This figure contrasts the controllable generation results of Prompt Diffusion with our improved Prompt Diffusion model on two novel, out-of-domain map-to-image tasks: Normal-to-Image and Canny-to-Image. The comparison aims to showcase the capabilities of both models in adapting to and performing on tasks they were not explicitly trained for.}
    \label{fig:map2img_ood}
\end{figure}

\subsection{Map2Img tasks}\label{sec:map2img}

In controllable generation, map-to-image tasks are crucial for evaluating a model's ability to reconstruct an image from one or more condition maps, encompassing elements such as line, shape, shade/light, and depth. Our initial evaluation of the models focuses on three specific map-to-image training tasks: depth-to-image, hed-to-image, and segmentation-to-image, all of which are in-domain tasks seen during training. Each of these tasks presents a typical challenge in controllable generation, a field in which a standard ControlNet often excels when it has been specifically trained for.

\paragraph{In-domain Map2Img tasks.} We first consider in-domain Map2Img tasks seen during training. 
In our qualitative evaluation, Figure~\ref{fig:map2img} presents the outcomes of controllable generation from several models: ControlNet, Prompt Diffusion, and two variants of improved Prompt Diffusion, where we conducted the generation process both with and without text prompts. Among all the evaluated models, the results produced by iPromptDiff-MG20M stand out as the most consistent and realistic. This is closely followed by the performance of iPromptDiff-IP2P. The other models, in comparison, tend to exhibit varying degrees of unnatural artifacts in their outputs.

In line with the evaluation protocol of Prompt Diffusion, we report the reconstruction Fr\'echet Inception Distance (FID) for our model. 
It's often presumed that a generalist model, trained across a broad range of tasks, may not be as effective as a specialist model when it comes to a particular in-domain task seen during training. However, as shown in Table~\ref{tab:fid}, our iPromptDiff model, despite its training as a multi-task generalist, achieves results that are on par with, or in some cases even slightly better than, those of task-specific models in controllable in-domain generation  tasks. This improvement in performance is attributable to our innovative method in learning and representing visual context, coupled with its effective fusion with linguistic guidance. This approach 
enables a more refined modulation of the visual context information within the textual guidance. This nuanced data is then processed by both the ControlNet and Stable Diffusion Components, ensuring an integrated and effective use of visual and textual information in the model's output.

\paragraph{Out-of-domain Map2Img tasks.}
In our exploration of out-of-domain Map-to-Image tasks that have not been seen during training, we have included results for both Normal-to-Image and Canny-to-Image in Figure~\ref{fig:map2img_ood}. In the Normal-to-Image task, both Prompt Diffusion and iPromptDiff demonstrate reasonable performance when provided with a full-text prompt. However, a notable difference emerges when the text prompt is set to an empty string (indicated with `n/a'). Under these conditions, Prompt Diffusion fails to produce coherent results, whereas iPromptDiff still manages to generate natural-looking images. This contrast effectively highlights the advantage of our model’s decoupled visual conditions and visual context-modulated text embedding. In scenarios where textual information is lacking or compromised, our model's reliance on visual context allows it to partly compensate for the absence of text guidance. A similar pattern is observed in the Canny-to-Image task, where, despite both methods producing below-average results, iPromptDiff maintains a certain level of consistency across different text prompt settings.

\begin{figure}[h!]
    \centering
    \includegraphics[width=0.45\textwidth]{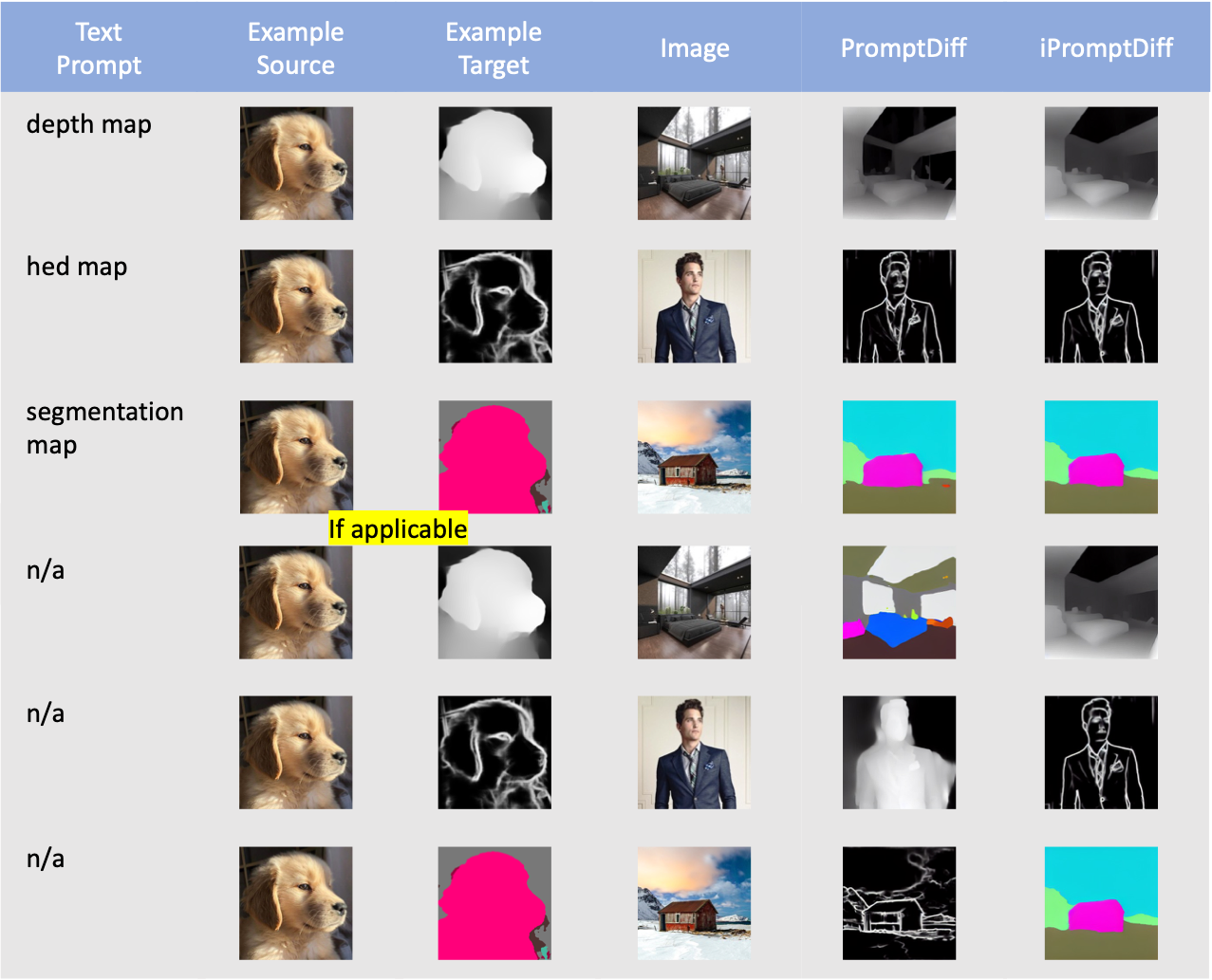}
    \caption{Comparison of map prediction results for in-domain tasks.}
    \label{fig:img2map}
\end{figure}

\subsection{Img2Map Tasks}

\paragraph{In-domain Img2Map tasks.}

The inverse of the map-to-image tasks discussed in Section \ref{sec:map2img}, such as image-to-hed, image-to-segmentation, and image-to-scribble (grouped under image-to-map tasks), typically present greater challenges. This increased difficulty arises because Stable Diffusion and other vision foundation models are typically not pretrained on condition map data of this nature. 
Another challenge pertains to the text encoder of the original Stable Diffusion. 

While using image captions as text prompts is straightforward in map-to-image tasks, the approach in image-to-map tasks requires careful consideration. Here, a complete image caption might be irrelevant to the task at hand, and an explicit task description could cause overfitting, thus impeding generalizability to unseen tasks. Adhering to the evaluation setup of Prompt Diffusion, we continue the practice of employing the task name as the text prompt for image-to-map tasks, whenever it is available. 

To explore potential text overfitting issues, we also assess the models without text prompts during testing. 
In this case, we designate the text prompt as an empty string, marked as `n/a'.
This forces the model to rely on in-context learning through provided example image pairs. We assess the models by calculating the root mean squared error (RMSE) between the actual condition maps of the query images and the corresponding maps generated by the models in response to these query images.

The results, presented in Table~\ref{tab:rmse-id}, show that when task names are explicitly given as text prompts (denoted as ``w/ text''), iPromptDiff demonstrates performance on par with ControlNet specialists and the Prompt Diffusion baseline in image-to-map tasks.  Conversely, when task names are masked with empty strings (denoted as ``w/o text''), our iPromptDiff demonstrates clear advantages over Prompt Diffusion.

\begin{table}[t!]
\centering
    \begin{adjustbox}{width=\linewidth}
    \begin{tabular}{@{} lccc @{}}
        \toprule
        & \multicolumn{3}{c}{w/ text} \\
        \midrule
        Model & Img-to-Depth & Img-to-HED & Img-to-Seg \\
        \midrule
        ControlNet & \textbf{0.2} & 0.18 & 0.36 \\
        PromptDiff & 0.21 & \textbf{0.16} & \textbf{0.32}  \\
        iPromptDiff (ours) & 0.22 & 0.17 & 0.34  \\
        iPromptDiff-MG20M (ours) & 0.21 & 0.19 & 0.34 \\
        \hhline{====}
        & \multicolumn{3}{c}{w/o text} \\
        \midrule
        Model & Img-to-Depth & Img-to-HED & Img-to-Seg \\
        \midrule
        PromptDiff & 0.34 & 0.25 & 0.41 \\
        iPromptDiff (ours) & 0.24 & 0.17 & \textbf{0.33} \\
        iPromptDiff-MG20M (ours) & \textbf{0.2} & \textbf{0.16} & \textbf{0.33} \\
        \bottomrule
    \end{tabular}
    \end{adjustbox}
    \caption{In-domain RMSE between test set condition maps and extracted maps by different models on the Instruct-Pix2Pix dataset.}
    \label{tab:rmse-id}
\end{table}

\paragraph{Out-of-domain Img2Map tasks.} We further conduct experiments on out-of-domain image-to-map tasks, including image-to-lineart and image-to-scribble, which were not included in the training phase. We use the line drawing processor~\cite{chan2022learning} and the PidiNet~\cite{su2021pixel} scribble processor to obtain the ground truth condition maps for image-to-lineart and image-to-scribble tasks respectively. Similar to in-domain tasks, we report the reconstruction RMSE in Table~\ref{tab:rmse-ood}. 

We again note that when both text prompts and visual in-context examples are provided during testing, iPromptDiff shows performance comparable to Prompt Diffusion. However, iPromptDiff significantly outperforms the baseline when the text prompt is absent. An additional intriguing finding is that for out-of-domain tasks, where text prompts are defaulted to the unseen task names, the iPromptDiff models tend to yield inferior results compared to the baseline. This may be attributed to the unique modality fusion process in iPromptDiff, as the concept implied by the unseen tasks' names is not introduced during training, leading to a mismatch between training and testing text prompts. This discrepancy can disrupt the vision-language interaction, potentially causing a degradation in performance. Interestingly, when the non-informative text prompts are entirely removed, iPromptDiff regains a significant advantage in out-of-domain generalizability.

\begin{table}[t!]
\centering
\begin{adjustbox}{width=\linewidth}
    \begin{tabular}{@{} lcc @{}}
        \toprule
        & \multicolumn{2}{c}{w/ text} \\
        \midrule
        Model & Img-to-Lineart & Img-to-Scribble\\
        \midrule
        PromptDiff & 0.\textbf{42} & \textbf{0.46} \\
        iPromptDiff (ours) & 0.45 & 0.47 \\
        iPromptDiff-MG20M (ours) & 0.5 & 0.53 \\
        \hhline{===}
        & \multicolumn{2}{c}{w/o text} \\
        \midrule
        Model & Img-to-Lineart & Img-to-Scribble \\
        \midrule
        PromptDiff & 0.35 & 0.35 \\
        iPromptDiff (ours) & 0.29 & 0.32 \\
        iPromptDiff-MG20M (ours) & \textbf{0.19} & \textbf{0.31} \\
        \bottomrule
    \end{tabular}
\end{adjustbox}
    \caption{Out-of-domain RMSE between test set condition maps and extracted maps by different models on Instruct-Pix2Pix dataset}
    \label{tab:rmse-ood}
    \vspace{-0.5em}
\end{table}

%% file: sec/5_discussion.tex
\section{Discussion and Future Work}
\label{sec:discussion}

\begin{figure}[t!]
    \centering
    \includegraphics[width=0.45\textwidth]{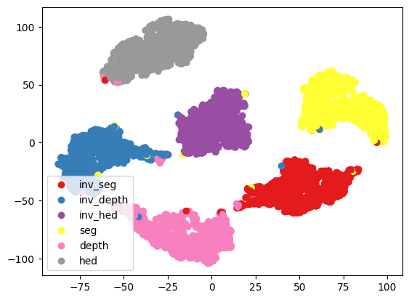}
    \caption{t-SNE visualization of the extracted visual context embeddings of test images.}
    \label{fig:tsne}
\end{figure}

\textbf{Visual context representation. }
As highlighted in Section~\ref{subsec:vision_enc}, the visual context extracted by the vision encoder plays a pivotal role in our proposed visual in-context learning framework.
 Through our experiments, we observed that the vision encoder in iPromptDiff effectively differentiates between clusters of various in-domain vision tasks post-training, as illustrated in Figure~\ref{fig:tsne}.
 This indicates the encoder's proficiency in discerning tasks based on provided example images. 
 It's important to note, however, that this capability is distinct from scenarios where explicit task descriptions are used as input conditions  \citep{wang2023incontext, qin2023unicontrol}.
The vision encoder in iPromptDiff undergoes end-to-end training, implicitly cultivating a more robust and generalizable representation of visual context. We believe that this aspect of our model underpins its potential adaptability to new, unseen tasks for which it has not been explicitly trained.

\textbf{Training task augmentation. }
A significant challenge in multitask training vision models lies in the limited number and diversity of vision tasks available. Key fundamental vision tasks like edge detection, depth estimation, and segmentation often lack either in quality, quantity, or variety, especially when compared to language tasks. To fully unlock the in-context learning ability of vision foundation models, it's essential to train them using a large-scale image dataset that encompasses a wide range of vision tasks. One practical approach to circumvent the constraint of limited vision tasks is through task augmentation. For instance, we can effectively increase the number of training tasks by mixing them up, $i.e.$,  merging the condition maps of an image into a novel map using diverse combination~weights. 

\textbf{Future work. }
While the current achievements of iPromptDiff are based on one-shot in-context examples, we note that the model can be readily adapted to accommodate multiple visual example pairs. This can be accomplished by averaging the embeddings of the visual context tokens. Alternatively, the perceiver resampler, as described by \citet{alayrac2022flamingo}, offers another solution. This approach is adept at encoding varying numbers of images into a consistent length of visual tokens, offering another viable pathway for enhancing the model’s capability to handle a broader range of visual inputs. Additionally, the selection of in-context examples represents a significant area for future exploration. The impact of choosing task-specific or query-specific examples on in-context learning (ICL) performance is well-documented in both language and vision domains, as evidenced by studies like \citet{liu2022makes}, \citet{zhang-etal-2022-active}, and \citet{ye2023compositional} in language, and \citet{zhang2023what} in vision. Incorporating a strategic example selection mechanism into our iPromptDiff model could lead to notable enhancements. This aspect presents a promising avenue for future research.

%% file: sec/6_conclusion.tex
\section{Conclusion}

In our paper, we tackle the crucial issue of visual understanding in visual in-context learning. We first discern the key difference between visual in-context examples and image queries, pinpointing a notable limitation in the existing Prompt Diffusion approach. Addressing this, we deploy a vision-transformer-based vision encoder that distinctly separates task comprehension from query processing, thereby refining the conditioning process. This encoder prepares an essential visual context representation to effectively modulate the text encoder, thus offering improved dual-mode (visual and linguistic) guidance for conditional generation. 
We anticipate key benefits from this model design, such as resilience to incomplete task descriptions and superior generalizability to out-of-domain tasks. These advantages are strongly corroborated by our experimental findings.

%% file: sec/X_suppl.tex
\clearpage
\setcounter{page}{1}
\maketitlesupplementary

\section{Training Details}
We train all of our model variants with $8\times$ Nvidia A100 for 10,000 steps ($\sim$ 2 days). The training data are processed into a minibatch for each training step. We utilized a per-card batch size of 64, culminating in a total batch size of 512. For efficiency, all images were preprocessed to a resolution of $256\times256$. Additionally, to bolster the models' classifier-free guidance capabilities, we implemented a strategy of randomly omitting text prompts, visual prompts (in-context examples), or both at varying rates: 0.4 for dropping text only, 0.4 for dropping visual examples only, and 0.1 for dropping both.

\section{CLIP Special Token and Visual-context Modulation}
We choose CLIP as the default text model for Stable Diffusion, whose vocabulary size is 49,408. We have modified the pretrained tokenizer to include a special token for our visual context learning, namely ``\_\_vc\_\_'', adding it to the existing vocabulary. This visual context token, once tokenized, is mapped to a vector representation extracted by our introduced vision encoder. During training, this special token is consistently prepended to the original text prompt. In this sense, it serves as a placeholder, which is later utilized for visual-context modulation. 

As mentioned in Section~\ref{subsec:vc_token}, simple concatenation is one naive way for implementing visual-context conditioning. However, we deem this approach sub-optimal, particularly when visual and linguistic information need to be processed together harmoniously to complement each other for complete image generation guidance. To substantiate this viewpoint, we conducted an ablative study comparing this basic method with our visual-context modulation technique. The quantitative findings are detailed in Table~\ref{tab:vc_comp}. In scenarios using full text prompts, our method matches performance in depth and HED map prediction tasks, while showing marked improvement in segmentation map prediction. Notably, when text prompts are removed, our method surpasses simple concatenation in all three map prediction tasks, clearly demonstrating the advantages of our visual-context modulation using the CLIP special token.

\begin{table}[h!]
    \begin{adjustbox}{width=0.45\textwidth}
    \begin{tabular}{lccc}
    \toprule
    & \multicolumn{3}{c}{w/ text} \\ 
    \midrule 
    Model & Img-to-Depth & Img-to-HED & Img-to-Seg \\
    \midrule
    iPromptDiff (concat.)  & 0.22  & 0.17  & 0.38  \\
    iPromptDiff (placeholder) & 0.22  & 0.17  & \textbf{0.34}  \\
    \hhline{====}
    & \multicolumn{3}{c}{w/ text} \\
    \midrule
    Model & Img-to-Depth & Img-to-HED & Img-to-Seg \\
    \midrule
    iPromptDiff (concat.)  & 0.25  & 0.27  & 0.39  \\
    iPromptDiff (placeholder) & \textbf{0.24}  & \textbf{0.17}  & \textbf{0.33}  \\
    \bottomrule
    \end{tabular}
    \end{adjustbox}
    \caption{Comparison between simple concatenation (abbreviated as ``concat.'') and visual context token placeholder}
    \label{tab:vc_comp}
\end{table}

\section{Effect of Positive Prompts}
Classifier-free guidance is commonly employed in diffusion models to improve output quality, where positive prompts like ``best quality'' and ``extremely detailed'' are involved. Typically, these prompts contribute positively to tasks like text-to-image generation and map-to-image controllable generation. However, in the context of map prediction, we have observed certain degree of negative impact on the performances of both Prompt Diffusion and iPromptDiff. This is partly illustrated in Figure~\ref{fig:a_prompt}, where we present actual generated images for a qualitative comparison. We see that Prompt Diffusion struggles to generate corresponding maps even with appropriate visual in-context examples, while iPromptDiff consistently makes accurate predictions of the right map types. This observation is further corroborated by our quantitative evaluation results. As indicated by the data in Table~\ref{tab:a_prompt}, the poorest performances are observed in scenarios where positive prompts are used exclusively as conditional prompts for classifier-free guidance. Again, iPromptDiff appears more robust than Prompt Diffusion, particularly because the former's performance does not change as much as the latter across all three scenarios.

\begin{figure*}[h!]
    \centering
    \includegraphics[width=0.7\linewidth]{figs/a\_prompt.png}
    \caption{Map prediction task outputs generated with positive prompts only.}
    \label{fig:a_prompt}
\end{figure*}

\begin{table}[h]
    \begin{adjustbox}{width=0.45\textwidth}
    \begin{tabular}{lccc}
    \toprule
    Model & Img-to-Depth & Img-to-HED & Img-to-Seg \\
    \midrule
    PromptDiff  & 0.21/0.42/0.34  & 0.16/0.36/0.25  & 0.32/0.46/0.41  \\
    iPromptDiff  & 0.22/0.23/0.24  & 0.17/0.26/0.17  & 0.34/0.35/0.33  \\
    \bottomrule
    \end{tabular}
    \end{adjustbox}
    \caption{\textbf{Negative impact of appending positive prompts on the RMSE of image-to-map tasks.} The experiment results of full text, positive prompts only and without text are separated by forward slash from left to right in order.}
    \label{tab:a_prompt}
\end{table}

\section{Higher-resolution Generation Results}

We provide higher-resolution generation results for figures in the main paper here. Specifically, Figure~\ref{fig:map2img_ood} corresponds to Figure~\ref{fig:map2img_ood_hq}, and Figure~\ref{fig:img2map} corresponds Figure~\ref{fig:img2map_hq}.

\begin{figure*}[h!]
    \centering
    \includegraphics[width=0.7\linewidth]{figs/map2img\_ood\_hq.png}
    \caption{Higher-resolution version of Figure~\ref{fig:map2img_ood}}
    \label{fig:map2img_ood_hq}
\end{figure*}

\begin{figure*}[h!]
    \centering
    \includegraphics[width=0.9\linewidth]{figs/img2map\_hq.png}
    \caption{Higher-resolution version of Figure~\ref{fig:img2map}}
    \label{fig:img2map_hq}
\end{figure*}